\documentclass[conference]{IEEEtran}
\usepackage{times}
\usepackage{epsfig}
\usepackage{graphicx}
\usepackage{amsmath}
\usepackage{amssymb}
\usepackage{color}

\usepackage{caption}
\usepackage{subcaption}
\ifCLASSINFOpdf
\else
\fi
\begin{document}
\title{Layer-Specific Adaptive Learning Rates \\for Deep Networks}

%\author{\IEEEauthorblockN{Bharat Singh, Soham De, Yangmuzi Zhang, Thomas Goldstein}
%\IEEEauthorblockA{School of Electrical and\\Computer Engineering\\
%University of Maryland\\
%College Park, Maryland 20742\\
%Email: bharat@cs.umd.edu}
%\and
%\IEEEauthorblockN{Homer Simpson}
%\IEEEauthorblockA{Twentieth Century Fox\\
%\and
%\IEEEauthorblockN{Homer Simpson}
%\IEEEauthorblockA{Twentieth Century Fox\\
%\and
%\IEEEauthorblockN{James Kirk\\ and Montgomery Scott}
%\IEEEauthorblockA{Starfleet Academy\\
%San Francisco, California 96678-2391\\


\author{\IEEEauthorblockN{Bharat Singh\IEEEauthorrefmark{1},
Soham De\IEEEauthorrefmark{1},
Yangmuzi Zhang\IEEEauthorrefmark{2}, 
Thomas Goldstein\IEEEauthorrefmark{1}, and
Gavin Taylor\IEEEauthorrefmark{3}}
\IEEEauthorblockA{\IEEEauthorrefmark{1}Department of Computer Science}
\IEEEauthorblockA{\IEEEauthorrefmark{2}Department of Electrical \& Computer Engineering\\
University of Maryland, College Park, MD 20742}
\IEEEauthorblockA{\IEEEauthorrefmark{3}Department of Computer Science, US Naval Academy, Annapolis, MD 21402\\
Email: \{bharat, sohamde, tomg\}@cs.umd.edu, ymzhang@umiacs.umd.edu,
taylor@usna.edu}}


\maketitle

\begin{abstract}
%\boldmath
The increasing complexity of deep learning architectures is resulting in
training time requiring weeks or even months. This slow training is due in
part to ``vanishing gradients,'' in which the gradients used by
back-propagation are extremely large for weights connecting deep layers
(layers near the output layer), and extremely small for shallow layers  (near
the input layer); this results in slow learning in the shallow layers.
Additionally, it has also been shown that in highly non-convex problems, such
as deep neural networks, there is a proliferation of high-error low curvature
saddle points, which slows down learning dramatically
\cite{pascanu2014saddle}. In this paper, we attempt to overcome the two above
problems by proposing an optimization method for training deep neural networks
which uses learning rates which are both specific to each layer in the network
and adaptive to the curvature of the function, increasing the learning rate at
low curvature points. This enables us to speed up learning in the shallow
layers of the network and quickly escape high-error low curvature saddle
points. We test our method on standard image classification datasets such as MNIST,
CIFAR10 and ImageNet, and demonstrate that our method increases accuracy as
well as reduces the required training time over standard algorithms.
\end{abstract}



% \fi
\IEEEpeerreviewmaketitle

\section{Introduction}
\label{sec:intro}

Deep neural networks have been extremely successful over the past few years,
achieving state of the art performance on a large number of tasks such as
image classification \cite{krizhevsky2012imagenet}, face recognition
\cite{taigman2014deepface}, sentiment analysis \cite{socher2013recursive},
speech recognition \cite{hinton2012deep}, etc. One can spot a general trend in
these papers: results tend to get better as the amount of training data
increases, along with an increase in the complexity of the deep network
architecture.  However, increasingly complex deep networks can take weeks or
months to train, even with high-performance hardware. Thus, there is a need
for more efficient methods for training deep networks.%, and this has been an
%active area of research recently.

Deep neural networks learn high-level features by performing a sequence of
non-linear transformations. Let our training data set $A$ be composed of $n$
data points $a_1, a_2, \dots, a_n \in \mathbb{R}^m$ and corresponding labels
$B=\{b_i\}_{i=1}^n.$ Let us consider a 3-layer network with activation
function $f$. Let $X_1$ and $X_2$ denote the weights on each layer that we are
trying to learn, i.e., $X_1$ denotes the weights between nodes of the first
layer and the second layer, and $X_2$ denotes the weights between nodes of the
second layer and the third layer. The learning problem for this specific
example can be formulated as the following optimization problem:
\begin{equation}
\label{eq:general_opt}
\underset{X_1, X_2}{\text{minimize}} \hspace{2mm} \big\| f(f(A \cdot X_1) \cdot X_2) - B \big\|_2^2
\end{equation}
The activation function $f$ can be any non-linear mapping, and is
traditionally a sigmoid or tanh function.  Recently, rectified linear (ReLu)
units ($f(z)=\max\{0, z\}$) have become popular because they tend to be
easy to train and yield superior results for some
problems~\cite{glorot2011deep}.

The non-convex objective (\ref{eq:general_opt}) is usually minimized using iterative methods (such as back-propagation) with the hope of converging to a good local minima.  Most iterative schemes generate additive updates to a set of parameters $x$ (in our case, the weight matrices) of the form
\begin{equation}
\label{eq:general_descent}
x^{(k+1)} = x^{(k)} + \Delta x^{(k)}
\end{equation}
where $\Delta x^{(k)}$ is some appropriately chosen update.
Notice we use slightly different notation here from standard optimization
literature in that we incorporate the step size or learning rate $t^{(k)}$
within $\Delta x^{(k)}$. This is done to help us describe other optimization
algorithms easily in the following sections. Thus, $\Delta x^{(k)}$ denotes
the update in the parameters, and comprises of a search direction and a step
size or learning rate $t^{(k)}$, which controls how large of a step to take in
that direction.

Most common update rules are variants of gradient descent, where the search direction is given by the negative gradient $g^{(k)}$:
\begin{equation}
\label{eq:sgd_param_update}
\Delta x^{(k)} = - t^{(k)} g^{(k)} = - t^{(k)} \nabla f(x^{(k)})
\end{equation}
Since the size of the training data for these deep networks is usually of the
order of millions or billions of data points, exact computation of the
gradient is not feasible. Rather, the gradient is often estimated using a
single data point or a small batch of data points. This is the basis for
stochastic gradient descent (SGD) \cite{robbins1951stochastic}, which is the
most widely used method for training deep nets. SGD requires manually
selecting an initial learning rate, and then designing an update rule for the
learning rate which decreases it over time (for example, exponential decay
with time). The performance of SGD, however, is very sensitive to this choice
of update, leading to \textit{adaptive} methods that automatically adjust the
learning rate as the system learns
\cite{zeiler2012adadelta,duchi2011adaptive}.

When these descent methods are used to train deep networks, additional
problems are introduced. As the number of layers in a network increases, the
gradients that are propagated back to the initial layers get very small. This
dramatically slows down the rate of learning in the initial layers, and
slows down convergence of the whole network \cite{hochreiter1997long}.

Recently, it has also been shown that for high-dimensional non-convex
problems, such as deep networks, the existence of local minima which have high
error relative to the global minima is exponentially small in the number of
dimensions. Instead, in these problems, there is an exponentially large number
of high error saddle points with low curvature \cite{pascanu2014saddle,
bray2007statistics, fyodorov2007replica}. Gradient descent methods, in
general, move away from saddle points by following the directions of negative
curvature. However, due to the low curvature of small negative eigenvalues,
the steps taken become very small, thus slowing down learning considerably.

In this paper, we propose a method that alleviates the problems mentioned above. The main contribution of our method is summarized below: 
\begin{itemize}
\item The learning rates are specific to each layer in the network. This
  allows larger learning rates to compensate for the small size of gradients
  in shallow layers.
\item The learning rates for each layer tend to increase at low curvature
  points. This enables the method to quickly escape from high-error,
  low-curvature saddle points, which occur in abundance in deep network.
\item It is applicable to most existing stochastic gradient optimization
  methods which use a global learning rate.
\item It requires very little extra computation over standard stochastic
  gradient methods, and requires no extra storage of previous gradients
  required as in AdaGrad \cite{duchi2011adaptive}.
\end{itemize}

In Section \ref{sec:related}, we review some popular gradient methods that
have been successful for deep networks. In Section \ref{sec:approach}, we
describe our optimization algorithm. Finally, in Section \ref{sec:results} we
compare our method to standard optimization algorithms on datasets like MNIST, CIFAR10
and ImageNet.

%!TEX root = main.tex

\section{Related Work}
\label{sec:related}
Stochastic Gradient Descent (SGD) still remains one of the most widely used
methods for large-scale machine learning, largely due to its ease in
implementation. In SGD, the updates for the parameters are defined by
equations (\ref{eq:general_descent}) and (\ref{eq:sgd_param_update}), and the
learning rate is decreased over time as iterates approach a local optimum. A
standard learning rate update is given by
\begin{equation}
\label{eq:global_learning_rate}
t^{(k)} = t^{(0)}/(1+\gamma k)^p
\end{equation}
where the initial learning rate $t^{(0)}$, $\gamma$ and $p$ are hyper-parameters chosen by the user.

Many modifications to the basic gradient descent algorithm have been proposed.
A popular method in the convex optimization literature is Newton's method,
which uses the Hessian of the objective function $f(x)$ to determine the step
size:
\begin{equation}
\label{eq:newton_step}
\Delta x^{(k)}_{nt} = - \nabla^2 f(x^{(k)})^{-1} g^{(k)}
\end{equation}

Unfortunately, as the number of parameters increases, even to moderate size,
computing the Hessian becomes very computationally expensive.  Thus, there
have been many modifications proposed which either try to improve the use of
first-order information or try to approximate the Hessian of the objective
function. In this paper, we focus on modifications to first-order methods.
%Thus, in the following sub-section, we review some of the more popular first-order methods proposed as modifications to the gradient descent algorithm, and that have been successfully used to train deep networks.

%\subsection{First Order Methods}
%\label{subsec:related_first_order}
%A number of first-order methods try to estimate a good learning rate based on the following heuristic: when the current estimate is far away from a local optimum, a higher learning rate should be used, whereas when the current estimate is near a local optimum, a lower learning rate should be used to ensure that the parameter values actually converge, and not oscillate near the optimum \cite{sutskever2013importance,duchi2011adaptive}.

The classical momentum method~\cite{polyak1964some} is a technique that
increases the learning rate for parameters for which the gradient consistently
points in the same direction, while decreasing the learning rate for
parameters for which the gradient is changing fast. Thus, the update equation
keeps track of previous updates to the parameters with an exponential decay:
\begin{equation}
\label{eq:classical_momentum}
\Delta x^{(k)} = \mu \Delta x^{(k-1)} - t g^{(k)}
\end{equation}
where $\mu \in [0,1]$ is called the momentum coefficient, and $t>0$ is the
global learning rate.

Nesterov's Accelerated Gradient (NAG)~\cite{nesterov1983method}, a first order
method, has a better convergence rate than gradient descent in certain
situations. This method predicts the gradient for the next iteration and
changes the learning rate for the current iteration based on the predicted
gradient. Thus, if the gradient is higher for the next step, it would increase
the learning rate for the current iteration and
if it is low, it would slow down. Recently, \cite{sutskever2013importance}
showed that this method can be thought of as a momentum method with the update
equation as follows:
\begin{equation}
\label{eq:nag_momentum}
\Delta x^{(k)} = \mu \Delta x^{(k-1)} - t \nabla f(x^{(k-1)} + \mu \Delta x^{(k-1)})
\end{equation}
Through a carefully designed random initialization and using a particular type
of slowly increasing schedule for $\mu$, this method can reach high levels of
performance when used on deep networks~\cite{sutskever2013importance}.

Rather than using a single learning rate over all parameters, recent work has
shown that using a learning rate specific to each parameter can be a much more
successful approach.  A method that has gained popularity is AdaGrad
\cite{duchi2011adaptive}, which uses the following update rule:
\begin{equation}
\label{eq:adagrad}
\Delta x^{(k)} = - \frac{t}{\sqrt{\sum_{i = 1}^k (g^{(i)})^2}} g^{(k)}
\end{equation}
The denominator is the $l_2$ norm of all the gradients of the previous
iterations. This scales the global learning rate $t,$ which is shared by all
the parameters, to give a parameter-specific learning rate. One disadvantage
of AdaGrad is that it accumulates the gradients over all previous iterations,
the sum of which continues to grow throughout training. This (along with weight decay)
shrinks the learning rate on each parameter until each is infinitesimally small, limiting
the number of iterations of useful training. 

A method which builds on AdaGrad and attempts to address some of the
above-mentioned disadvantages is AdaDelta~\cite{zeiler2012adadelta}. AdaDelta
accumulates the gradients in the previous time steps using an exponentially
decaying average of the squared gradients. This prevents the denominator from
becoming infinitesimally small, and ensures that the parameters continue to be
updated even after a large number of iterations. It also replaces the global
learning rate $t$ with an exponentially decaying average of the squares of the
parameter updates $\Delta x$ over the previous iterations. This method has
been shown to perform relatively well when used to train deep networks, and is
much less sensitive to the choice of hyper-parameters. However, it does not
perform as well as other methods like SGD and AdaGrad in terms of
accuracy \cite{zeiler2012adadelta}.

\section{Our Approach}
\label{sec:approach}

Because of the ``vanishing gradients'' phenomenon, shallow network layers tend
to have much smaller gradients than deep layers -- sometimes differing by
orders of magnitude from one layer to the next~\cite{hochreiter1997long}.  In
most previous work in optimization for deep networks, methods either keep a
global learning rate that is shared over all parameters, or use an adaptive
learning rate specific to each parameter. Our method exploits the following
observation: {\em parameters in the same layer have gradients of similar
magnitudes, and can thus efficiently share a common learning rate.}
Layer-specific learning rates can be used to accelerate layers with smaller
gradients. Another advantage of this approach is that by avoiding the
computation of large numbers of parameter-specific learning rates, our method
remains computationally efficient. Finally, as mentioned in Section
\ref{sec:intro}, to avoid slowing down learning at high-error low curvature
saddle points, we also want our method to take large steps at low curvature
points.

Let $t^{(k)}$ be the learning rate at the $k$-th iteration for any standard
optimization method. In case of SGD, this would be given by equation
\ref{eq:global_learning_rate}, while for AdaGrad it would just be the global
learning rate $t$ as in equation \ref{eq:adagrad}. We propose to modify
$t^{(k)}$ as follows:
\begin{equation}
\label{eq:layer_gradient}
t^{(k)}_l = t^{(k)} (1 + \log (1+ 1/(\| g^{(k)}_l \|_2)))
\end{equation}
Here $t^{(k)}_l$ denotes the new learning rate for the parameters in the
$l$-th layer at the $k$-th iteration and $g^{(k)}_l$ denotes a vector of the
gradients of the parameters in the $l$-th layer at the $k$-th iteration. Thus,
we see that we use only the gradients in the same layer to determine the
learning rate for that layer. It is also important to note that we do not use
any gradients from previous iterations, and thus save on storage.

From equation \ref{eq:layer_gradient}, we see that when the gradients in a layer are very large, the equation just reduces to using the normal learning rate $t^{(k)}$. However, when the gradients are very small, we are more likely to be near a low curvature point. Thus, the equation scales up the learning rate to ensure that the initial layers of the network learn faster, and that we escape high-error low curvature saddle points quickly.

We can use this layer-specific learning rate on top of SGD. Using equation \ref{eq:sgd_param_update}, the update in that case, would be:
\begin{eqnarray}
\label{eq:sgd_layer}
\Delta x^{(k)}_l &=& - t^{(k)}_l g^{(k)}_l \\
&=& - t^{(k)} (1 + \log (1+ 1/(\| g^{(k)}_l \|_2))) g^{(k)}_l
\end{eqnarray}
where $\Delta x^{(k)}_l$ denotes the update in the parameters of the $l$-th layer at the $k$-th iteration.

\begin{table*}[t]
\centering
\begin{tabular}{|c|c|c|c|c|c|c|}
\hline
\textbf{Iteration} & \textbf{SGD} & \textbf{Ours-SGD} & \textbf{Nesterov} & \textbf{Ours-NAG} & \textbf{AdaGrad} & \textbf{Ours-AdaGrad}\\ \hline
200 & 7.90 $\pm$ 0.44 & 7.25 $\pm$ 0.46 & 6.66 $\pm$ 0.47 & 5.37 $\pm$ 0.5 & 4.12 $\pm$ 0.32 & 3.40 $\pm$ 0.3 \\ \hline
600 & 3.29 $\pm$ 0.22 & 3.05 $\pm$ 0.21 & 3.01 $\pm$ 0.19 & 2.84 $\pm$ 0.17 & 2.21 $\pm$ 0.14 & 1.95 $\pm$ 0.16 \\ \hline
1000 & 1.89 $\pm$ 0.08 & 1.80 $\pm$ 0.13 & 1.92 $\pm$ 0.07 & 1.83 $\pm$ 0.18 & 1.68 $\pm$ 0.11 & 1.57 $\pm$ 0.14 \\ \hline
1400 & 1.60 $\pm$ 0.11 & 1.49 $\pm$ 0.09 & 1.74 $\pm$ 0.12 & 1.52 $\pm$ 0.11 & 1.61 $\pm$ 0.08 & 1.61 $\pm$ 0.09 \\ \hline
1800 & 1.52 $\pm$ 0.09 & 1.41 $\pm$ 0.12 & 1.56 $\pm$ 0.09 & 1.37 $\pm$ 0.09 & 1.41 $\pm$ 0.09 & 1.34 $\pm$ 0.08 \\ \hline
\end{tabular}
\caption{Mean error rate on MNIST after different iterations for stochastic gradient descent, Nesterov's accelerated gradient and AdaGrad with their layer specific adaptive versions are shown in the table. Each method was run 10 times and their mean and standard deviation is reported.}
\label{tab:stat_mnist}
\end{table*}

Similarly, we can modify AdaGrad's update equation \eqref{eq:adagrad} to use our modified learning rates.
\begin{equation}
\label{eq:adagrad_layer}
\Delta x^{(k)}_l = - \frac{t^{(k)}_l}{\sqrt{\sum_{i = 1}^k (g^{(i)}_l)^2}} g^{(k)}_l
\end{equation}
Note that, unlike AdaGrad which uses a distinct learning rate for each
parameter, we use a different learning rate for each {\em layer}, which is
shared by all weights in that layer. Additionally, AdaGrad modifies the
learning rate based on the entire history of gradients observed for that
weight while we update a layer's learning rate based only on gradients
observed for all weights in a specific layer in the {\em current} iteration.
Thus, our scheme avoids both storing gradient information from previous
iterations and computing learning rates for each parameter; it is therefore
less computationally and memory intensive when compared to AdaGrad. The
proposed layer specific learning rates also works well on large scale datasets
like ImageNet (when applied over SGD), where AdaGrad fails to converge to a
good solution.

The proposed method can be used with any existing optimization technique
which uses a global learning rate, provides a layer-specific learning rate,
and escapes saddle points quickly, all without sacrificing computation or
memory usage. As we show in Section \ref{sec:results}, using our adaptive
learning rates on top of existing optimization techniques almost always
improves performance on standard datasets.

The proposed method can be used with any existing optimization technique
which uses a global learning rate. This helps in getting a layer-specific
learning rate, as well as, helps in escaping saddle points quicker, with very
little computational overhead. As we show in Section \ref{sec:results}, using
our adaptive learning rates on top of existing optimization techniques almost
always improves performance on standard datasets.

%!TEX root = main.tex

\setlength{\belowcaptionskip}{-0.1pt}

\section{Experimental Results}
\label{sec:results}

\begin{figure*}[t]
\centering
        \begin{subfigure}[b]{0.33\textwidth}
                \centering
                \includegraphics[width=\linewidth]{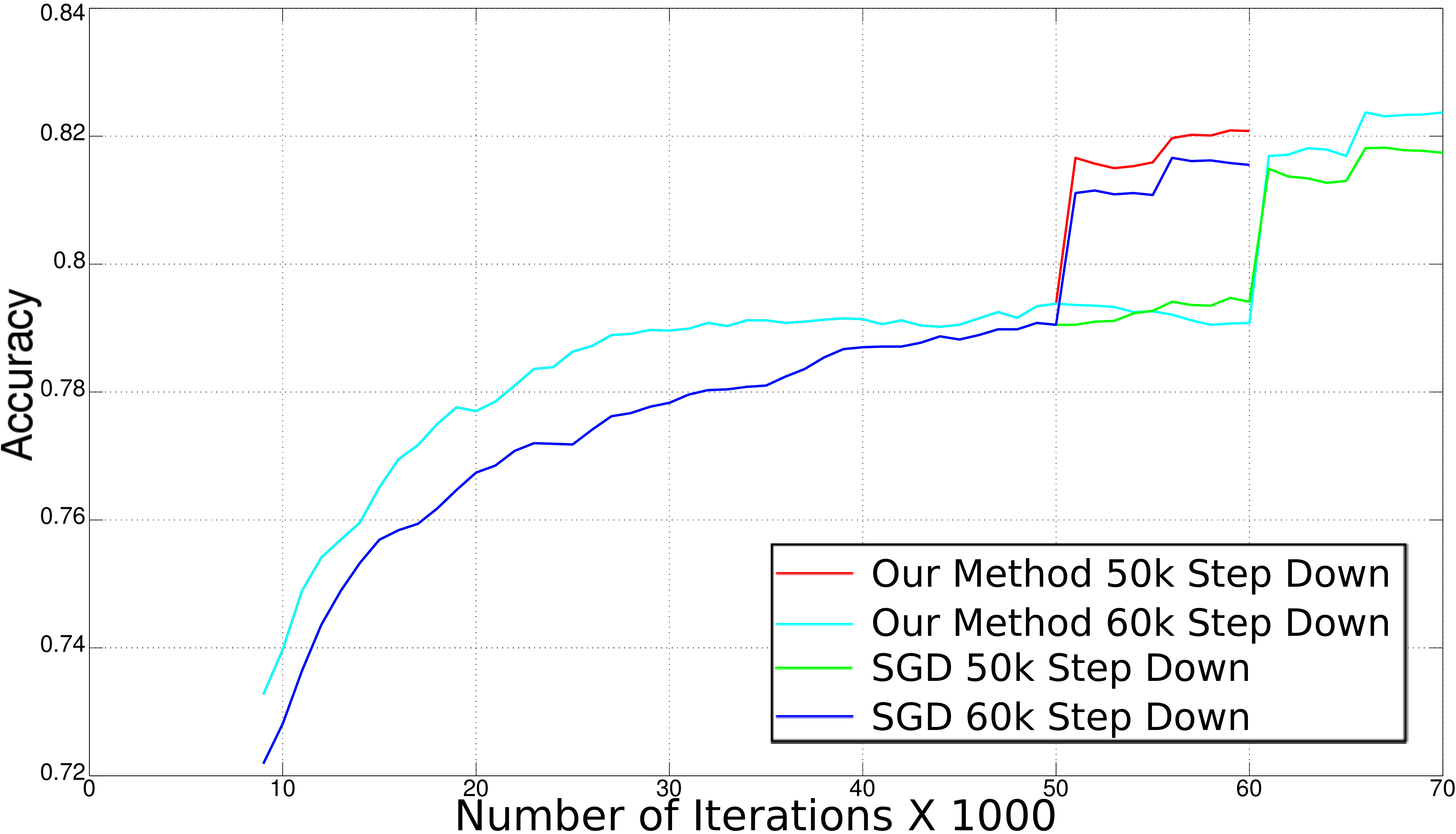}
                \caption{Stochastic Gradient Descent}
                \label{fig:cifar_acc_sgd}
        \end{subfigure}\hfill
        \begin{subfigure}[b]{0.33\textwidth}
                \centering
                \includegraphics[width=\linewidth]{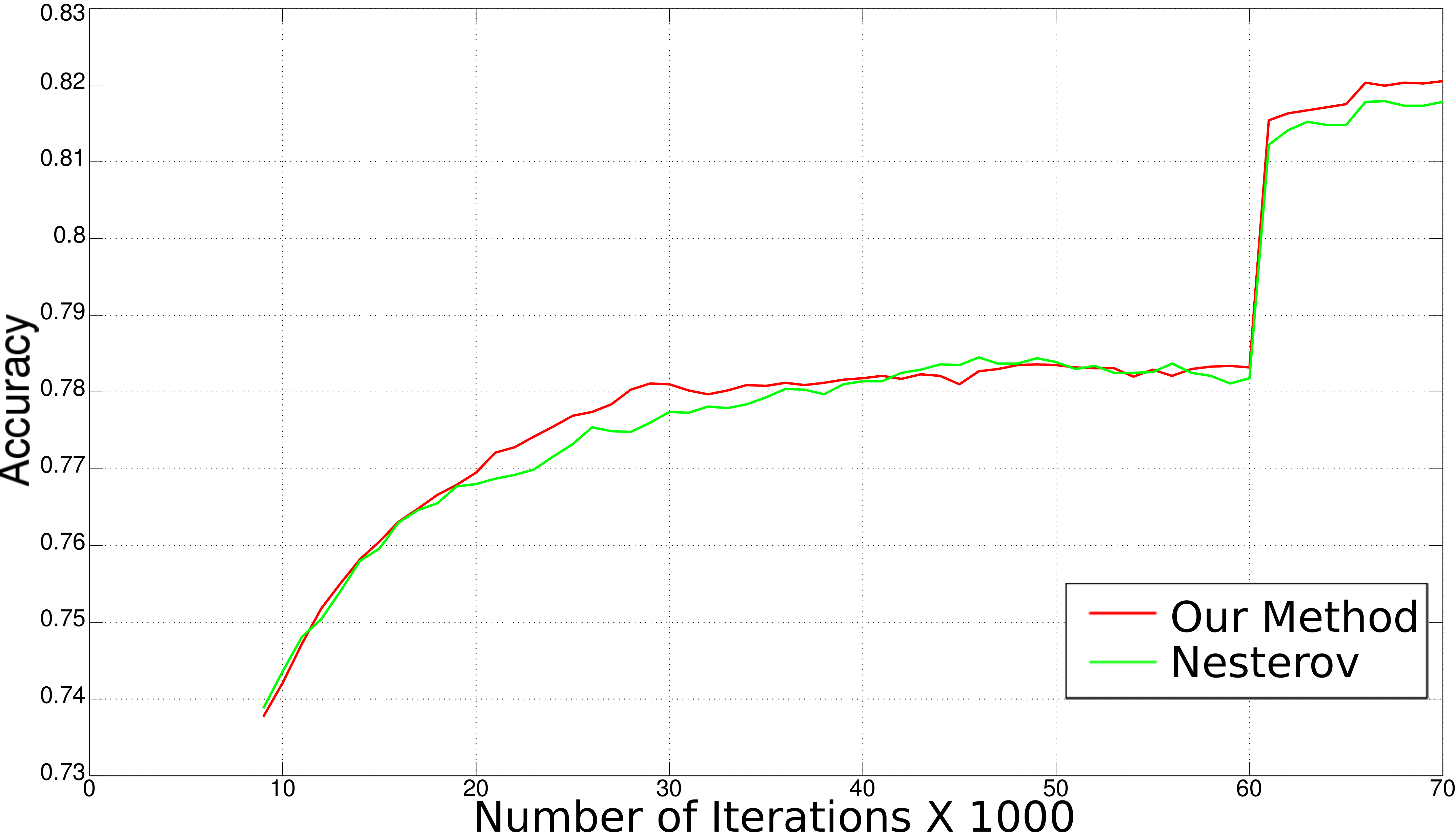}
                \caption{Nesterov's Accelerated Gradient}
                \label{fig:cifar_acc_nest}
        \end{subfigure}\hfill
        \begin{subfigure}[b]{0.33\textwidth}
                \centering
                \includegraphics[width=\linewidth]{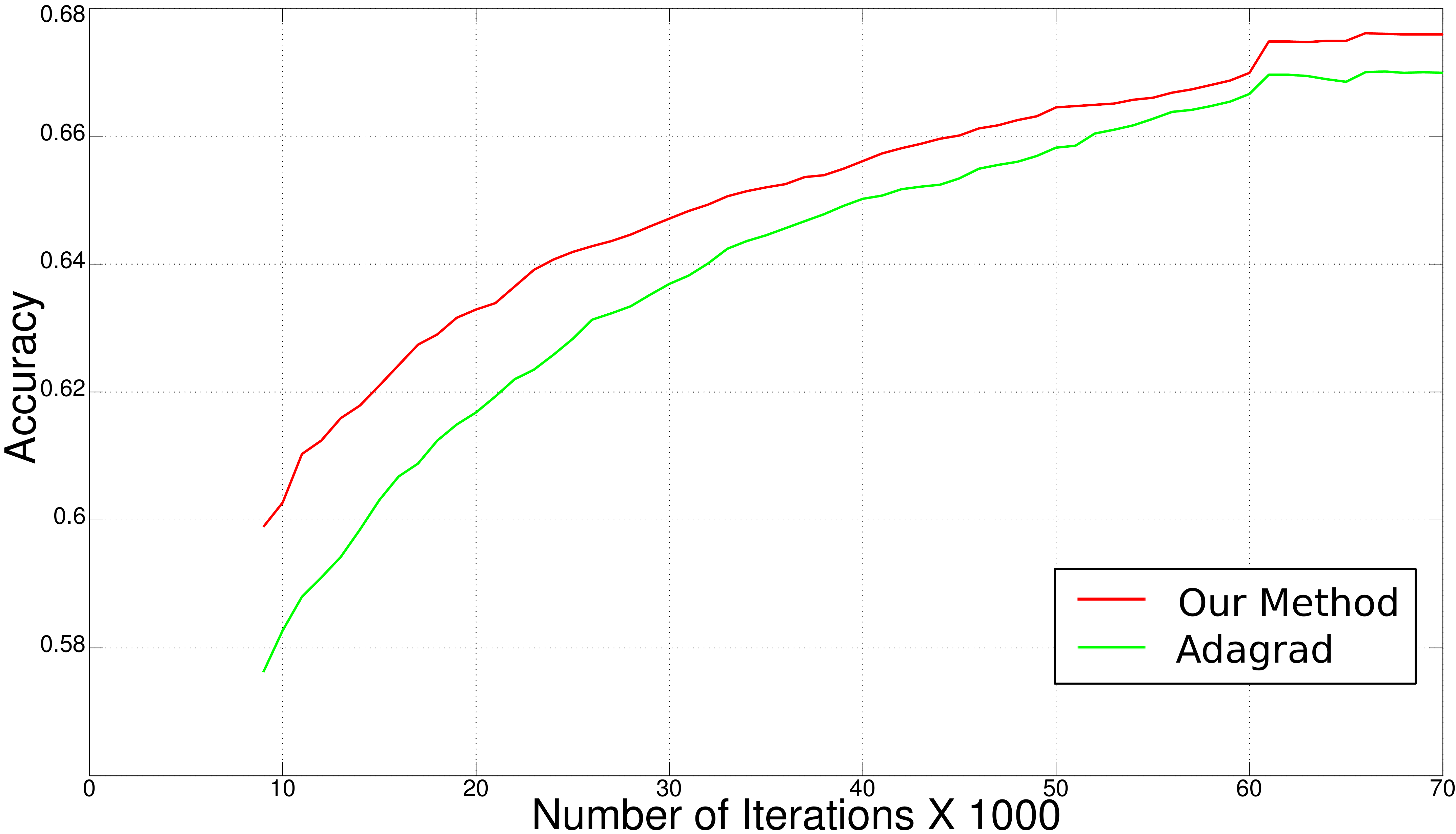}
                \caption{AdaGrad}
                \label{fig:cifar_acc_adagrad}
        \end{subfigure}
        \caption{On CIFAR data set: plots showing accuracies
      (Figures \ref{fig:cifar_acc_sgd}-\ref{fig:cifar_acc_adagrad}) comparing
    SGD, NAG and AdaGrad, each with our adaptive layer-wise learning rates.
  For the SGD plot, we show results both when we step down the learning rate
at 50,000 iterations as well as 60,000 iterations.}
        \label{fig:cifar_plots}
\end{figure*}
\subsection{Dataset}
We present image classification results on three standard datasets: MNIST, CIFAR10 and
ImageNet (ILSVRC 2012 dataset, part of the ImageNet challenge). MNIST contains 60,000 handwritten digit images for training and 10,000 handwritten digit
images for testing. CIFAR10 contains has 10 classes with 6,000 images in
each class. ImageNet contains 1.2 million color images from 1000 different
classes. 

\subsection{Experimental Details}
We use Caffe \cite{jia2014caffe} to implement our method. Caffe provides
optimization methods for Stochastic Gradient Descent (SGD), Nesterov's
Accelerated Gradient (NAG) and AdaGrad. For a fair comparison between
state-of-the-art methods, we add our adaptive layer-specific learning rate
method on top of each of these optimization methods. In our experiments, we
demonstrate the effectiveness of our algorithm on convolutional neural
networks on 3 datasets. On CIFAR10,  we use the same global learning rate as
provided in Caffe. Since our method always increases the layer-specific
learning rate (with respect to other optimization methods) based on the
global learning rate, we start with a slightly smaller learning rate of 0.006
to make the learning less aggressive for the ImageNet experiment. SGD was
initialized with the learning rate used in \cite{krizhevsky2012imagenet} for
experiments done on ImageNet.

\subsubsection{MNIST}
We use the same architecture as LeNet for our experiments on MNIST.  We present the results of using our proposed layer-specific learning rates on top of stochastic gradient descent, Nesterov's accelerated gradient method and AdaGrad on the MNIST dataset. Since all methods converge very quickly on this dataset, we present the accuracy and loss only for the first 2,000 iterations. Table \ref{tab:stat_mnist} shows the mean accuracy and standard deviation when each method was run 10 times. We observe that our proposed layer-specific learning rate is consistently better than Nesterov's accelerated gradient, stochastic gradient descent and AdaGrad. In all the experiments, the proposed method also attains the maximum accuracy of 99.2\% just like stochastic gradient descent, Nesterov's accelerated gradient and AdaGrad. 

\subsubsection{CIFAR10}
On CIFAR10 we use a convolutional neural network with 2 layers of 32 feature maps from $5\times5$ convolution kernels, each followed by $3\times3$ max pooling layers. After this we have another convolution layer with 64 feature maps from a $5\times5$ convolution kernel followed by a $3\times3$ max pooling layer. Finally, we have a fully connected layer with 10 hidden nodes and a soft-max logistic regression layer. After each convolution layer a ReLu non-linearity is applied. This is the same architecture as specified in Caffe. For the first 60,000 iterations the learning rate was 0.001 and it was dropped by a factor of 10 at 60,000 and 65,000 iterations.

\begin{table*}
\centering
%\resizebox{\columnwidth}{!}{
\begin{tabular}{|c|c|c|c|c|c|c|}
\hline
\textbf{Iteration} & \textbf{SGD} & \textbf{Ours-SGD} & \textbf{Nesterov} & \textbf{Ours-NAG} & \textbf{AdaGrad} & \textbf{Ours-AdaGrad}\\ \hline
5000 & 68.8 $\pm$ 0.49 & 70.10 $\pm$ 0.89 & 69.36 $\pm$ 0.31 & 70.10 $\pm$ 0.59 & 54.90 $\pm$ 0.26 & 57.53 $\pm$ 0.67 \\ \hline
10000 & 74.05 $\pm$ 0.51 & 74.48 $\pm$ 0.59 & 73.17 $\pm$ 0.25 & 74.00 $\pm$ 0.29 & 58.26 $\pm$ 0.58 & 60.95 $\pm$ 0.59 \\ \hline
25000 & 77.40 $\pm$ 0.32 & 77.43 $\pm$ 0.15 & 76.17 $\pm$ 0.61 & 77.29 $\pm$ 0.59 & 63.02 $\pm$ 0.95 & 64.90 $\pm$ 0.57 \\ \hline
60000 & 78.76 $\pm$ 0.87 & 78.74 $\pm$ 0.38 & 78.35 $\pm$ 0.33 & 78.18 $\pm$ 0.65 & 66.86 $\pm$ 0.93 & 68.03 $\pm$ 0.23 \\ \hline
70000 & 81.78 $\pm$ 0.14 & 82.10 $\pm$ 0.32 & 81.75 $\pm$ 0.25 & 81.92 $\pm$ 0.26 & 67.04 $\pm$ 0.91 & 68.30 $\pm$ 0.39 \\ \hline
\end{tabular}
%}
\caption{Mean accuracy on CIFAR10 after different iterations for SGD, NAG and AdaGrad with their layer specific adaptive versions are shown in the table. The mean and standard deviation over 5 runs is reported.}
\label{tab:stat_cifar}
\end{table*}

On this dataset, we again observe that final error and loss of our method is consistently lower than SGD, NAG and AdaGrad (Table \ref{tab:stat_cifar}). After step down, our adaptive method reaches a lower accuracy than both SGD and NAG. Note that just using our optimization method (without changing the network architecture) we can get an improvement of 0.32\% over the mean accuracy for SGD. Even if we step down the learning rate at 50,000 iterations (taking 60000 iterations in total), we obtain an accuracy of 82.08\%, which is better than SGD after 70,000 iterations, significantly cutting down on required training time Fig. \ref{fig:cifar_plots}. Since our method converges much faster when used with SGD, it is possible to perform the step down on the learning rate even earlier, potentially reducing training time even further. Although Adagrad does not perform very well on CIFAR10 with default parameters, we observe an improvement of 1.3\% over the mean final accuracy, with again a significant speed-up in training time.

\subsubsection{ImageNet}
We use an implementation of AlexNet \cite{krizhevsky2012imagenet} in Caffe, a deep convolutional neural network architecture, for comparing our method with other optimization algorithms. AlexNet consists of 5 convolution layers followed by 3 fully connected layers. More details regarding the architecture can be found in the paper \cite{krizhevsky2012imagenet}. 

\begin{figure}[t]
\centering
        \begin{subfigure}[b]{0.45\textwidth}
                \centering
                \includegraphics[width=\linewidth]{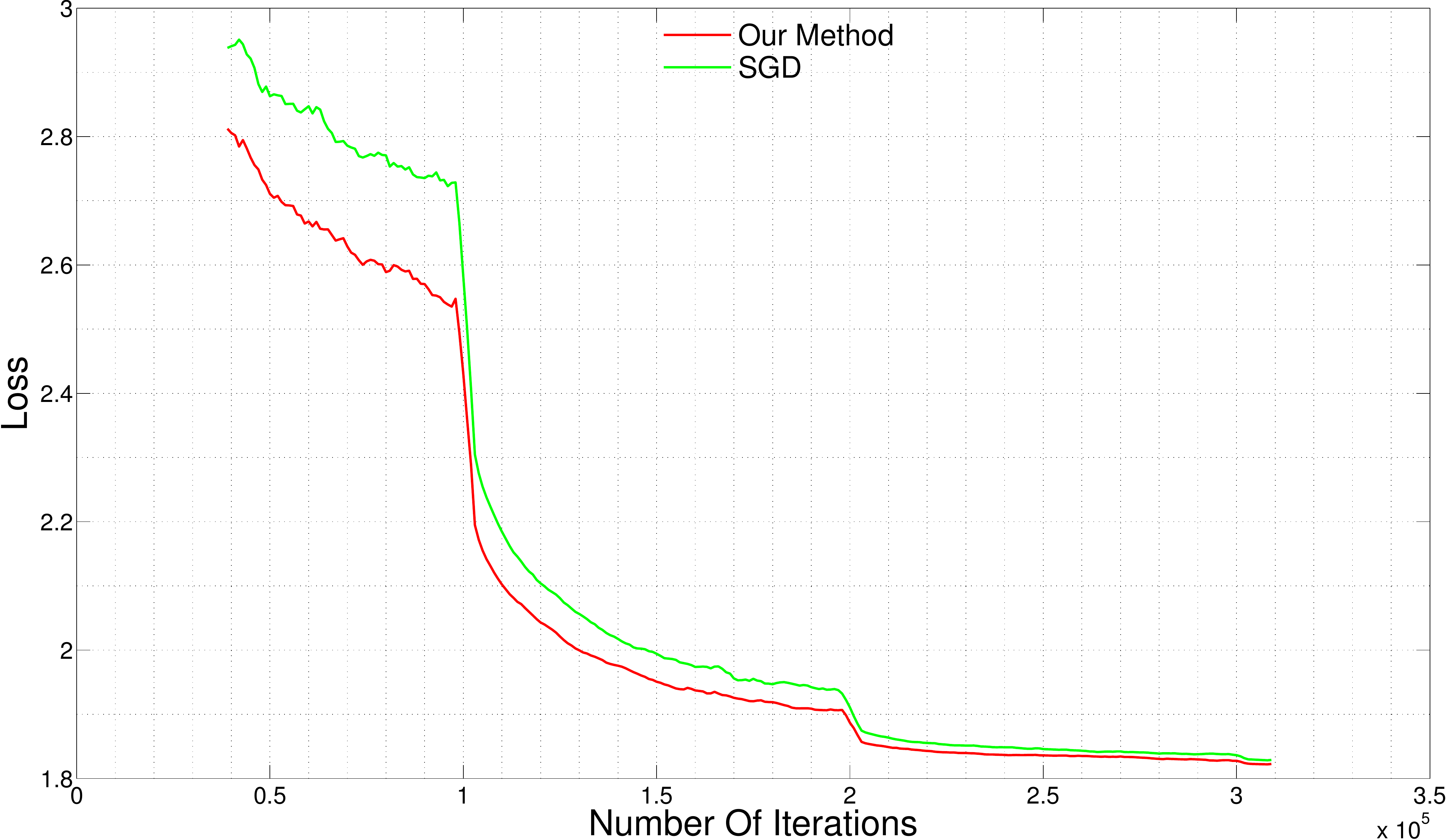}
                \caption{Loss with Stochastic Gradient Descent}
                \label{fig:imagenet_loss_sgd}
        \end{subfigure}
        \begin{subfigure}[b]{0.45\textwidth}
                \centering
                \includegraphics[width=\linewidth]{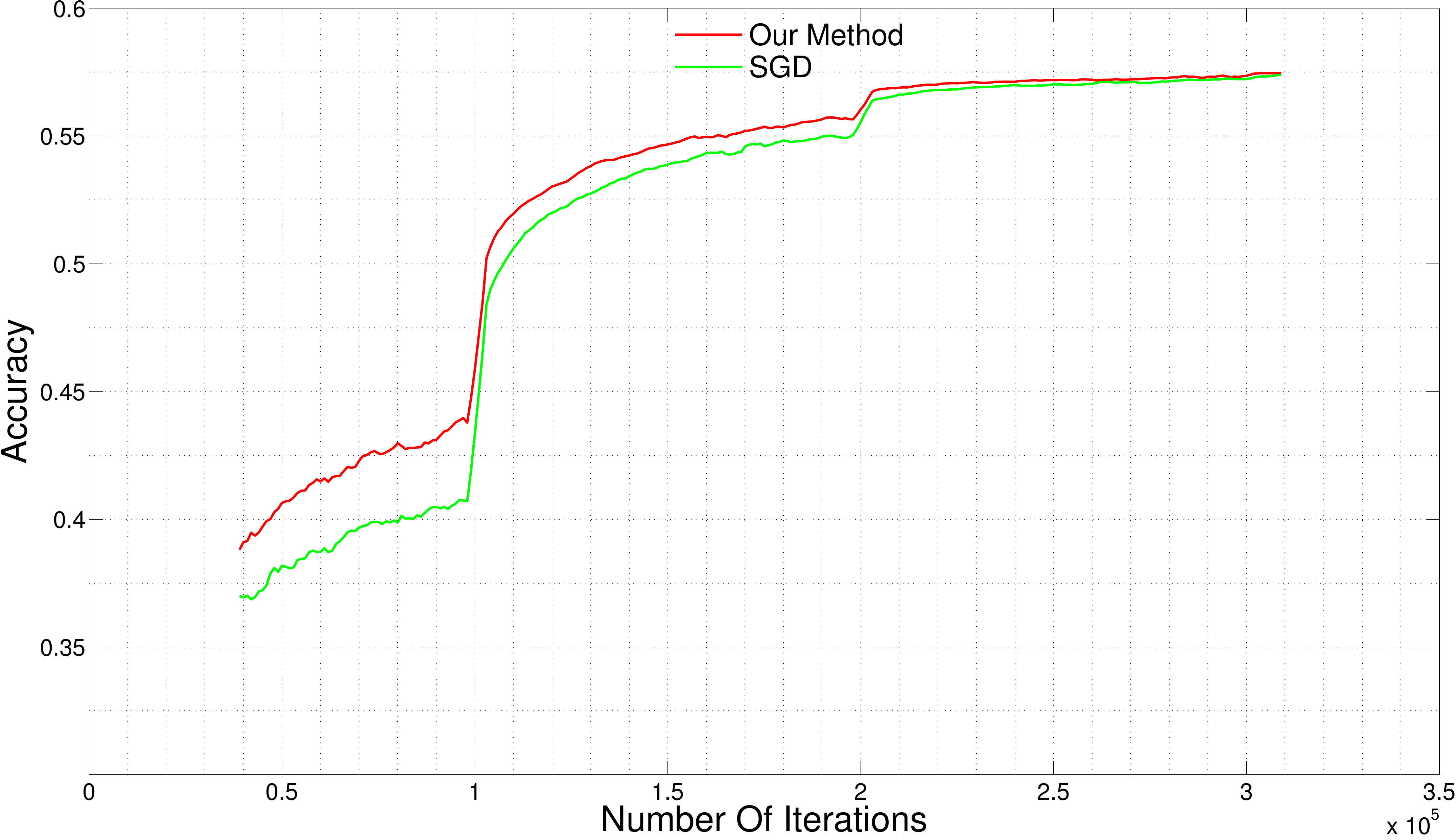}
                \caption{Accuracy with Stochastic Gradient Descent}
                \label{fig:imagenet_acc_sgd}
        \end{subfigure}
        \caption{On ImageNet data set: plot comparing stochastic gradient descent with our adaptive layer-wise learning rates. We can see a consistent improvement in accuracy and loss over the regular SGD method across all iterations.}
        \label{fig:imagenet_plots}
\end{figure}

Since AlexNet is a deep neural network with significant complexity, it is suitable to apply our method on this network architecture. Fig \ref{fig:imagenet_plots} shows the results of using our method over SGD. We observe that our method obtains significantly greater accuracy and lower loss after 100,000 and 200,000 iterations. Further, we are also able to reach the maximum accuracy of 57.5\% on the validation set after 295,000 iterations which is achieved by SGD only after 345,000 iterations, resulting in a reduction of 15\% in training time. Given that such a large model takes more than a week to train properly, this is a significant reduction. Our loss is also consistently lower than SGD across all iterations. In the existing model, we perform a step down by a factor of 10 after every 100,000 iterations. In order to analyze how our method performs when we reduce the number of training iterations, we vary the number of training iterations at a specific learning rate before performing a step down. Table \ref{tab:step_down} shows the final accuracy after 350,000 iterations of SGD and our method. Although the final accuracy drops slightly as we decrease the number of iterations after which we perform the step down in the learning rate, it is clearly evident that our method achieves better accuracy than SGD. Note that we report top-1 class accuracy. Since we use the Caffe implementation of the AlexNet architecture and do not use any data augmentation techniques, our results are slightly lower than those reported in \cite{krizhevsky2012imagenet}.

\begin{table}
\centering
\begin{tabular}{|c|c|c|}
\hline
\textbf{Iterations} & \textbf{SGD} & \textbf{Our Method} \\ \hline
70,000 & 55.72\% & 55.84\% \\ \hline
80,000 & 56.25\% & 56.57\% \\ \hline
90,000 & 56.96\% & 57.13\% \\ \hline 
\end{tabular}
\caption{Comparison of stochastic gradient descent and Our Method with step-down at different iterations on ImageNet}
\label{tab:step_down}
\end{table}

%!TEX root = main.tex

\section{Conclusions}
\label{sec:conclusions}

In this paper we propose a general method for training deep neural networks
using layer-specific adaptive learning rates, which can be used on top of any
optimization method with a global learning rate. The method uses gradients
from each layer to compute an adaptive learning rate for each layer. It aims
to speed up convergence when the parameters are in a low curvature saddle
point region.  Layer-specific learning rates also enable the method to prevent
slow learning in initial layers of the deep network, usually caused by very
small gradient values.

%To the best of our knowledge, this is the first work that proposes learning
%rates being adapted on layer-specific information. In the context of deep
%networks, it makes a lot of sense intuitively to think about different layers
%having different learning rates. We plan to study this in more detail in our
%future work. We also plan to look at more efficient methods of quickly
%escaping high-error low curvature saddle points, which occur in abundance in
%the highly non-convex neural network problems.


\section*{Acknowledgment}
The authors acknowledge ONR Grant numbers N0001415WX01341 and N000141512676, as well as the University of
Maryland supercomputing resources (http://www.it.umd.edu/hpcc).



%\bibliographystyle{IEEEtran}

\bibliographystyle{IEEEtran}
%\bibliography{IEEEabrv,references}
%\nocite{*}
\bibliography{references}

\begin{thebibliography}{10}
\providecommand{\url}[1]{#1}
\csname url@samestyle\endcsname
\providecommand{\newblock}{\relax}
\providecommand{\bibinfo}[2]{#2}
\providecommand{\BIBentrySTDinterwordspacing}{\spaceskip=0pt\relax}
\providecommand{\BIBentryALTinterwordstretchfactor}{4}
\providecommand{\BIBentryALTinterwordspacing}{\spaceskip=\fontdimen2\font plus
\BIBentryALTinterwordstretchfactor\fontdimen3\font minus
  \fontdimen4\font\relax}
\providecommand{\BIBforeignlanguage}[2]{{%
\expandafter\ifx\csname l@#1\endcsname\relax
\typeout{** WARNING: IEEEtran.bst: No hyphenation pattern has been}%
\typeout{** loaded for the language `#1'. Using the pattern for}%
\typeout{** the default language instead.}%
\else
\language=\csname l@#1\endcsname
\fi
#2}}
\providecommand{\BIBdecl}{\relax}
\BIBdecl

\bibitem{pascanu2014saddle}
R.~Pascanu, Y.~N. Dauphin, S.~Ganguli, and Y.~Bengio, ``On the saddle point
  problem for non-convex optimization,'' \emph{arXiv preprint arXiv:1405.4604},
  2014.

\bibitem{krizhevsky2012imagenet}
A.~Krizhevsky, I.~Sutskever, and G.~E. Hinton, ``Imagenet classification with
  deep convolutional neural networks,'' in \emph{Advances in neural information
  processing systems}, 2012, pp. 1097--1105.

\bibitem{taigman2014deepface}
Y.~Taigman, M.~Yang, M.~Ranzato, and L.~Wolf, ``Deepface: Closing the gap to
  human-level performance in face verification,'' in \emph{Computer Vision and
  Pattern Recognition (CVPR), 2014 IEEE Conference on}.\hskip 1em plus 0.5em
  minus 0.4em\relax IEEE, 2014, pp. 1701--1708.

\bibitem{socher2013recursive}
R.~Socher, A.~Perelygin, J.~Y. Wu, J.~Chuang, C.~D. Manning, A.~Y. Ng, and
  C.~Potts, ``Recursive deep models for semantic compositionality over a
  sentiment treebank,'' in \emph{Proceedings of the Conference on Empirical
  Methods in Natural Language Processing (EMNLP)}.\hskip 1em plus 0.5em minus
  0.4em\relax Citeseer, 2013, pp. 1631--1642.

\bibitem{hinton2012deep}
G.~Hinton, L.~Deng, D.~Yu, G.~E. Dahl, A.-r. Mohamed, N.~Jaitly, A.~Senior,
  V.~Vanhoucke, P.~Nguyen, T.~N. Sainath \emph{et~al.}, ``Deep neural networks
  for acoustic modeling in speech recognition: The shared views of four
  research groups,'' \emph{Signal Processing Magazine, IEEE}, vol.~29, no.~6,
  pp. 82--97, 2012.

\bibitem{glorot2011deep}
X.~Glorot, A.~Bordes, and Y.~Bengio, ``Deep sparse rectifier networks,'' in
  \emph{Proceedings of the 14th International Conference on Artificial
  Intelligence and Statistics. JMLR W\&CP Volume}, vol.~15, 2011, pp. 315--323.

\bibitem{robbins1951stochastic}
H.~Robbins, S.~Monro \emph{et~al.}, ``A stochastic approximation method,''
  \emph{The Annals of Mathematical Statistics}, vol.~22, no.~3, pp. 400--407,
  1951.

\bibitem{zeiler2012adadelta}
M.~D. Zeiler, ``Adadelta: An adaptive learning rate method,'' \emph{arXiv
  preprint arXiv:1212.5701}, 2012.

\bibitem{duchi2011adaptive}
J.~Duchi, E.~Hazan, and Y.~Singer, ``Adaptive subgradient methods for online
  learning and stochastic optimization,'' \emph{The Journal of Machine Learning
  Research}, vol.~12, pp. 2121--2159, 2011.

\bibitem{hochreiter1997long}
S.~Hochreiter and J.~Schmidhuber, ``Long short-term memory,'' \emph{Neural
  computation}, vol.~9, no.~8, pp. 1735--1780, 1997.

\bibitem{bray2007statistics}
A.~J. Bray and D.~S. Dean, ``Statistics of critical points of gaussian fields
  on large-dimensional spaces,'' \emph{Physical review letters}, vol.~98,
  no.~15, p. 150201, 2007.

\bibitem{fyodorov2007replica}
Y.~V. Fyodorov and I.~Williams, ``Replica symmetry breaking condition exposed
  by random matrix calculation of landscape complexity,'' \emph{Journal of
  Statistical Physics}, vol. 129, no. 5-6, pp. 1081--1116, 2007.

\bibitem{polyak1964some}
B.~T. Polyak, ``Some methods of speeding up the convergence of iteration
  methods,'' \emph{USSR Computational Mathematics and Mathematical Physics},
  vol.~4, no.~5, pp. 1--17, 1964.

\bibitem{nesterov1983method}
Y.~Nesterov, ``A method of solving a convex programming problem with
  convergence rate o (1/k2),'' in \emph{Soviet Mathematics Doklady}, vol.~27,
  no.~2, 1983, pp. 372--376.

\bibitem{sutskever2013importance}
I.~Sutskever, J.~Martens, G.~Dahl, and G.~Hinton, ``On the importance of
  initialization and momentum in deep learning,'' in \emph{Proceedings of the
  30th International Conference on Machine Learning (ICML-13)}, 2013, pp.
  1139--1147.

\bibitem{jia2014caffe}
Y.~Jia, E.~Shelhamer, J.~Donahue, S.~Karayev, J.~Long, R.~Girshick,
  S.~Guadarrama, and T.~Darrell, ``Caffe: Convolutional architecture for fast
  feature embedding,'' \emph{arXiv preprint arXiv:1408.5093}, 2014.

\end{thebibliography}

%\begin{thebibliography}{1}


%\end{thebibliography}

\end{document}